%% file: main.tex
\documentclass[11pt]{article}

\usepackage{microtype} 
\usepackage{booktabs}  
\usepackage{url}  
\usepackage{paralist}

\usepackage{amsmath}
\usepackage{amsthm}

\usepackage{graphicx}

\usepackage[final]{automl}

\usepackage{natbib}

\title{Are encoders able to learn landmarkers for warm-starting of Hyperparameter Optimization?}

\author[1]{\nameemail{Antoni Zajko}{antoni.zajko.stud@pw.edu.pl}}
\author[1]{\nameemail{Katarzyna Woźnica}{katarzyna.woznica@pw.edu.pl}}

\affil[1]{Warsaw University of Technology}

\hypersetup{%
  pdfauthor={}, 
  pdftitle={},
  pdfsubject={},
  pdfkeywords={}
}

\begin{document}

\maketitle

\begin{abstract}
Effectively representing heterogeneous tabular datasets for meta-learning purposes is still an open problem. Previous approaches rely on representations that are intended to be universal. This paper proposes two novel methods for tabular representation learning tailored to a specific meta-task -- warm-starting Bayesian Hyperparameter Optimization. Both follow the specific requirement formulated by ourselves that enforces representations to capture the properties of landmarkers. The first approach involves deep metric learning, while the second one is based on landmarkers reconstruction. We evaluate the proposed encoders in two ways. Next to the gain in the target meta-task, we also use the degree of fulfillment of the proposed requirement as the evaluation metric. Experiments demonstrate that while the proposed encoders can effectively learn representations aligned with landmarkers, they may not directly translate to significant performance gains in the meta-task of HPO warm-starting.
\end{abstract}


\input{sections/introduction}
\input{sections/overview}
\input{sections/representation}
\input{sections/experiments}
\input{sections/discussion}

\begin{acknowledgements}
Research was funded by Warsaw University of Technology within the Excellence Initiative: Research University (IDUB) programme.
\end{acknowledgements}

\bibliographystyle{abbrvnat}
\bibliography{main}

\input{sections/checklist}

\newpage
\appendix

\input{sections/appendix}

\end{document}

%% file: sections/introduction.tex
\section{Introduction}
Tabular data is the most common type of data, which is ubiquitous across various domains of science and industry~\citep{davenport_potential_2019, alanazi_using_2022}. Such data has properties that pose unique challenges when working with it. One is inherent heterogeneity, which involves varying dimensionality, types of feature and target variables, and their distributions. This heterogeneity makes it challenging to compare two datasets and limits the use of meta-learning methods in this field. So far, existing approaches to data representation either rely on handcrafted meta-features~\citep{wistuba_mf1} or are based on meta-models that are believed to grasp the internal structure of the data~\citep{dataset2vec, kim2018learningwarmstartbayesianhyperparameter}. One of the solutions to that problem is representation learning, which aims to learn such meta-features from the data itself. Of particular interest are the encoders, neural networks intended to learn the intrinsic properties of data, which can be used for similarity comparison of datasets.

Among the applications of the tabular dataset representation is Bayesian Hyperparameter Optimization (Bayesian HPO)~\citep{bo_hpo} meta-task solving the hyperparameter tuning problem. It aims to search for the best hyperparameter configuration of the machine learning model using an iterative algorithm, which is proven to outperform Random Search~\citep{turner2021bayesianoptimizationsuperiorrandom}. However, Bayesian HPO has its limitations. One of them is a cold-start problem~\citep{bai2023transferlearningbayesianoptimization}, meaning that before the proper optimization phase, the algorithm needs to perform several random trials to gain initial information on the hyperparameter space. To mitigate that, several approaches are proposed. One of them is providing a set of the initial points, called warm-start points, that are expected to improve either the convergence speed or final result of the procedure~\citep{bai2023transferlearningbayesianoptimization}. The promising approach to searching such configurations is using meta-learning methods, i.e., using knowledge from the previous datasets that are believed to be similar to target one~\citep{feurer_mf2}. For this class of solutions, further methods of calculating the similarity of the datasets are needed. One of the directions with the highest potential is representation learning~\citep{metabu, dataset2vec}.

There are some encoder-based solutions~\citep{kim2018learningwarmstartbayesianhyperparameter, dataset2vec} for the problem of the warm-staring Bayesian HPO, but several factors bias their evaluation. Firstly, the solutions to the Bayesian HPO are mainly based on the selection from a predefined set of hyperparameters' configurations (portfolio). However, the baselines they are compared to are often based on random sampling from the entire hyperparameter space, yielding too optimistic comparative results. Additionally, evaluations that only use HPO can be biased by the noise introduced by exploring the hyperparameter space after the initial phase of the optimization procedure. Therefore, there is a need for additional ways of evaluation of such encoders. 

\paragraph{Problem motivation} It is shown that the representations that are intended to be general may not be sufficient for warm-starting Bayesian HPO~\citep{rethinking}. Therefore, there is a need for a tabular data representation that is aware of this meta-task. This research aims to provide a representation tailored to the specific meta-task -- warm-starting Bayesian HPO. Such a representation is expected to contain relevant information for this meta-task, providing significant performance gains. In this context, it means that warm-start configurations found using such representations will lead to either a speed-up in convergence or a more optimal final point at the end of the optimization procedure.

\paragraph{Requirement}\label{ass:land_align} To achieve that, the proposed encoder can be created with a view of the requirement that corresponds to the target meta-task. Namely, we propose a following requirement:

\textit{
Distances between representations of the datasets should align with the distances between corresponding landmarkers.
}

Such an assumption is promising due to the fact that landmarkers themselves contain information about the internal structure of data in relation to the performance's variability across hyperparameter configurations. However, in practical application, we can not use them due to their unacceptably high computational cost. Hence, creating encoders that can reflect information carried by the landmarkers can be a reasonable step.

\paragraph{Key contributions} In this work:
\begin{enumerate}
    \item We propose two novel approaches to learning representations of tabular datasets that are specific to the meta-task of the warm-starting Bayesian HPO. The first approach is a representation based on a deep metric learning that aims to reflect distances in the landmarkers space. The second approach is a representation based on the reconstruction of landmarkers. This architecture consists of an encoder finding a latent representation of the dataset, and in the next step, this latent representation is used to predict the landmarker vector. The reconstructed landmarker vector is the obtained representation of the dataset here.
    \item We provide a detailed analysis of obtained representations using two evaluation methods. The first is a well-established method based on the performance in warm-starting Bayesian HPO~\citep{dataset2vec}. For this evaluation, we propose various baselines that only measure the performance of the selection algorithm -- random selection and simple heuristic. The proposed second one can capture whether the produced representations align with the landmarkers.
\end{enumerate}

%% file: sections/overview.tex
\section{Overview}
\begin{figure}[h]
    \centering
    \includegraphics[width=\linewidth]{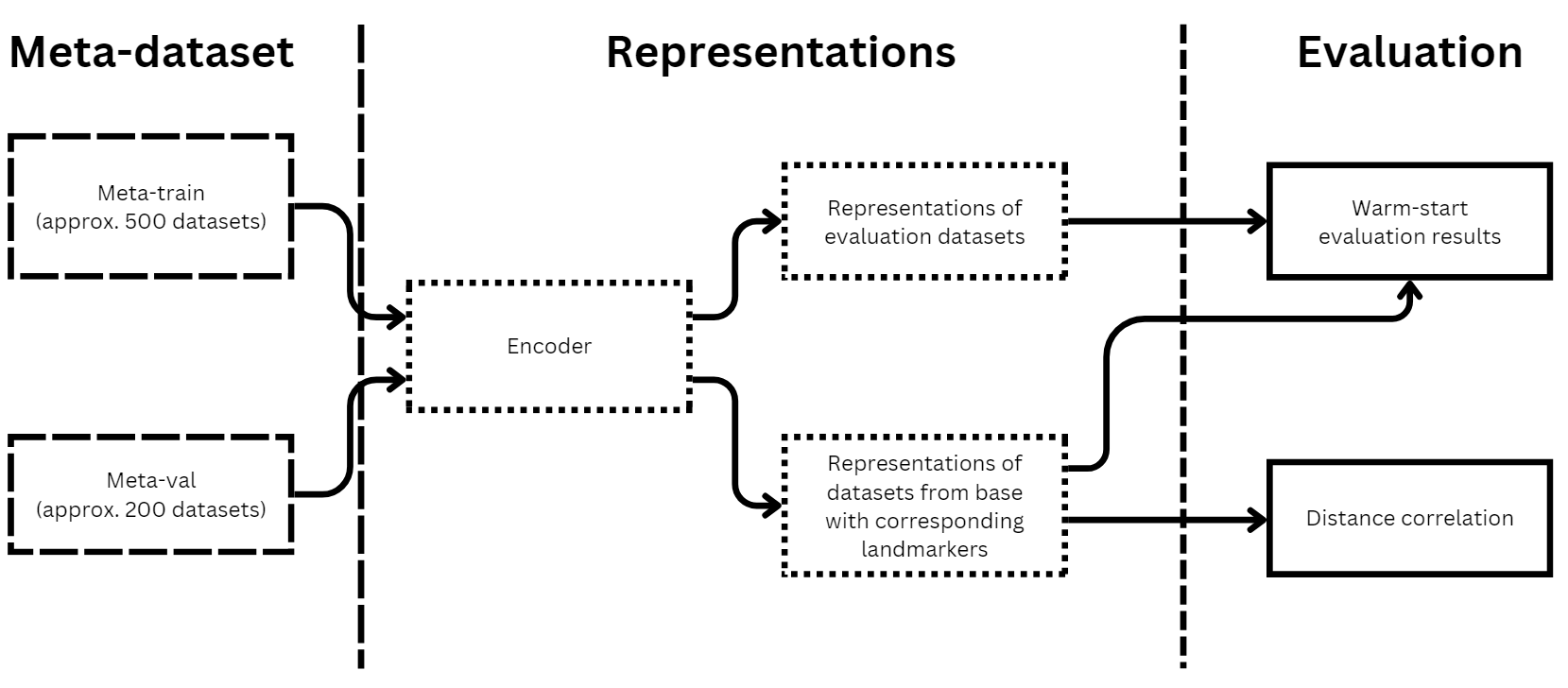}
    \caption{Visualization of the relation between data, encoders, and results. The dashed lines correspond to datasets, the dotted line corresponds to intermediate results, and the solid line depicts what is evaluated in the proposed methods.}
    \label{fig:problem_setting}
\end{figure}

In this section, we present an overview of the methodology used to design and evaluate the proposed encoders. We describe the meta-dataset used and then introduce the encoder creation and evaluation process. Their graphical view is presented in Figure~\ref{fig:problem_setting}.

For all experiments, we use a meta-dataset consisting of various tabular datasets with binary classification problems. This set of datasets is divided into two meta-samples -- meta-train and meta-validation. Meta-train is used for two purposes:
\begin{inparaenum}[(1)]
    \item for training of considered encoders,
    \item for building the base of datasets' landmarkers, which is used for warm-starting Bayesian HPO. In this context, the sample serves two purposes --it is used for the selection of configurations for the portfolio and as data to the base of precomputed landmarkers.
\end{inparaenum}
On the other hand, meta-validation serves as the set of data on which we calculate metrics to assess the quality of the encoders. The detailed data preparation procedure description is in Section~\ref{sec:metadataset}.

Having meta-training and meta-validation sets, we train three encoders, which consist of Dataset2Vec from~\citep{dataset2vec} serving as a baseline and two approaches we proposed. They are described in Section~\ref{sec:representation}. Afterwards, we employ the obtained encoders to calculate the representation of the datasets for both the meta-training and meta-validation samples. With all prepared assets, we can validate the encoders. Obtained encoders are evaluated in two ways. The first way aims at the validation of \hyperref[ass:land_align]{Requirement} and is to calculate how distances between obtained representations align with distances of corresponding landmarkers. The second approach of encoder assessment is the evaluation of possible gains in the Bayesian HPO using Average Distance To Minimum (ADTM)~\citep{wistuba_mf1} and Critical Distance (CD) plot~\citep{IsmailFawaz2018deep}.

%% file: sections/representation.tex
\section{Representation}
\label{sec:representation}
This section describes the encoder-based approaches to the problem of tabular dataset representation -- one established in~\citep{dataset2vec} and two proposed by ourselves. The first one uses the deep metric learning approach~\citep{Kaya_2019,kim2018learningwarmstartbayesianhyperparameter,mohan2023deepmetriclearningcomputer}. The second approach predicts the landmarkers vector based on the representation of the dataset, i.e., produces representation that is expected to be landmarkers.

\subsection{Dataset2Vec}
\label{sec:dataset2vec}
All proposed approaches are based on Dataset2Vec mapping datasets into latent space, which is assumed to have desired spatial properties~\citep{dataset2vec}. It is a DeepSet-based method~\citep{zaheer_deep_2017} composed of feed-forward neural networks. Dataset2Vec extracts interdependencies inside the data between features, targets and joint distributions in a three-step process, which results in fixed-size encoding. According to the authors, it meets several requirements, including scalability and expressivity, which in practical applications mean that these representations can be used in a wide range of meta-tasks. However, \citet{rethinking} shows that such requirements are not necessarily sufficient for the Bayesian HPO warm-starting meta-task.

\subsection{Representations based on the deep metric learning}
\label{sec:dataset2vec_metric}
\begin{figure}[h]
    \centering
    \includegraphics[width=0.8\linewidth]{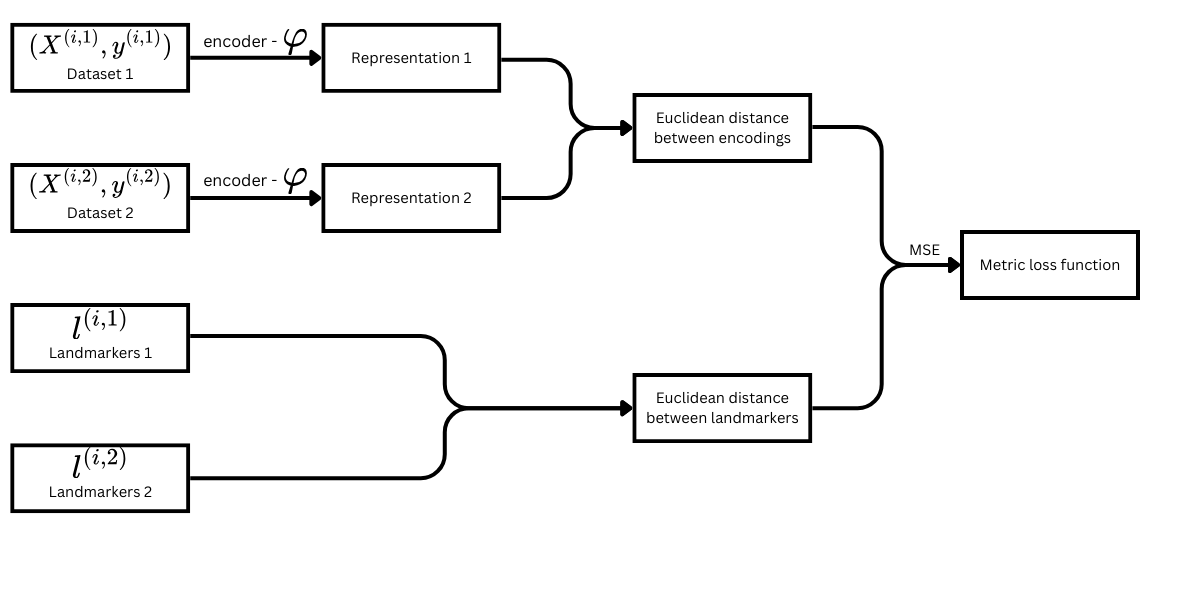}
    \caption{Schematic view of the Equation~\ref{eq:metric_loss}, which calculates the Mean Squared distance between Euclidean distances of meta-features and landmarkers. The first component (top) is the distance between feature representations, and the second (bottom) is the distance between landmarkers.}
    \label{fig:metric_loss}
\end{figure}

To address the issue of Dataset2Vec's meta-features that are insufficient for warm-starting Bayesian HPO meta-task, we propose an approach that follows the \hyperref[ass:land_align]{Requirement} and is based on metric learning. The origin of this approach is similar to one presented by~\citet{kim2018learningwarmstartbayesianhyperparameter}. It aims to train an encoder to project datasets to the latent space in which the Euclidean distances of corresponding vectors align with the distances between landmarkers. The encoder is the same as in~\citep{dataset2vec}, but we change the loss function. The proposed function is expressed as
\begin{equation}
    loss_{metric} = \frac{1}{N}\sum_{i=1}^{N}d\left(\varphi(X^{(i, 1)}, y^{(i, 1)}), \varphi(X^{(i, 2)}, y^{(i, 2)})\right)^2 - d(l^{(i, 1)}, l^{(i, 2)})^2,
    \label{eq:metric_loss}
\end{equation} 
 and consists of two parts. The first component is an Euclidean distance between representations created by the encoder of the datasets. The $\varphi$ denotes and encoder, $(X^{(i, \cdot)}, y^{(i, \cdot)})$ the dataset consisting of feature matrix $X$ and target vector $y$ and $i$ is an index of the pair of datasets and the second index indicates the number of the dataset in pair -- either 1 or 2. The second component is Euclidean distance between landmarkers, which are denoted by $l^{(i, \cdot)}$ (indexing the same as before). Having these two components, the formula evaluates the Mean Squared Error (MSE) between them (see Figure~\ref{fig:metric_loss}).

\subsection{Representations based on the landmarkers reconstruction}
\label{sec:dataset2vec_reconstruction}
The second method is based on a different way of satisfying the \hyperref[ass:land_align]{Requirement}. It creates dataset representation by prediction of the landmarkers. So, the encoder consists of two components: Dataset2Vec and reconstruction network -- a multi-layer perceptron. During the forward propagation of the dataset, it is first passed through the Dataset2Vec denoted by $\varphi$ to produce meaningful latent representations. Afterward, it is passed to the reconstruction component denoted by $\rho$ to obtain the output vector, predicted landmarkers of dimensionality $p$, which are treated as a final dataset representation. In the training of this two-step encoder, as the loss function, we treat an MSE between predicted and true landmarkers expressed as 

\begin{equation}
    loss_{reconstruction} = \frac{1}{N}\sum_{n=1}^{N} \frac{1}{p}\sum_{k=1}^{p}(\rho(\varphi(X^{(i)}, y^{(i)}))_k - l_k)^2.
    \label{eq:reconstruction_loss}
\end{equation}


%% file: sections/experiments.tex
\section{Experiments}
In this section, we first describe the meta-dataset. Next, we introduce baselines. Finally, we provide a detailed description of the evaluation process and show the results. The code for reproduction is available under URL \href{https://anonymous.4open.science/r/wsmf-32C5/README.md}{https://anonymous.4open.science/r/wsmf-32C5/README.md}. The computation takes around 10 hours on a single machine with RTX4060 GPU, AMD Ryzen 7 7840HS CPU, and 32 GB of RAM.

\subsection{Meta-dataset}
\label{sec:metadataset}
We download approximately 700 datasets from OpenML~\citep{vanschoren_openml_2013} to create a meta-dataset. We choose datasets with at most 50 features, 100000 observations, and 10 classes. Next, we remove columns that seem to be IDs. Additionally, we binarize target variables to obtain binary classification tasks that are as balanced as possible. All categorical columns are one-hot encoded, and numerical columns are scaled to $[0, 1]$ interval. Next, within each dataset, we perform a train-test split. The meta-dataset prepared in this way is split into two samples - meta-train and meta-validation. The first contains around $500$ datasets, while the second has $200$.

In this research, we consider only the xgboost algorithm~\citep{xgboost} due to its competitive performance in machine learning on tabular data~\citep{probst_tunability_nodate, Bent_jac_2020}. Due to that, by portfolio, we mean a set of configurations of the xgboost algorithm, and landmarkers of the dataset denote the performances of these configurations on this data. The grid of the hyperparameters is provided in Appendix~\ref{app:hp_grid}.

\subsection{Landmarkers}
The core of meta-training and evaluation are landmarkers vectors. One of the most important parts of calculating them is a selection of the evaluated configurations. We choose them based only on the meta-train sample. The is performed in three steps. Firstly, we calculate simple meta-features, e.g., the mean of all numerical columns, the number of categorical columns, the prevalence of positive class, etc. Next, we cluster datasets based on the computed meta-features using K-Means algorithm~\citep{lloyd_kmeans}. The number of clusters equals the number of configurations in the portfolio, which, in our case, is $100$. Finally, we select configurations for the portfolio using these clusters. To achieve that, we perform tournament selection where each cluster is treated as a tournament, and the winner is the best on average configuration on all datasets in a cluster. 

\subsection{Evaluation}
\label{sec:metrics}
We evaluate the obtained encoders in two ways. The first method focuses on assessing landmarkers properties reflection of obtained outputs, while the second assesses their usability in warm-starting Bayesian HPO.

\subsubsection{Distance correlation}
\label{sec:distance_corr}
The primary is based on the evaluation of the correlation of distances of encoder-based representations compared to landmarkers, which results in the assessment of the degree of the fulfillment of the \hyperref[ass:land_align]{Requirement}. To measure that, we use Spearman correlation~\citep{zar2005spearman} (further denoted by $corr$) between Euclidean distances of outputs of encoders to each other and distances of landmarkers corresponding to considered datasets. In both cases, we conduct evaluation on $S=20$ samples out of $N=1000$ pairs of datasets. For approaches that do not produce predictions of the landmarkers, i. e. Dataset2Vec and metric learning approach, we report correlations of the distances between representations of the datasets to distances between corresponding landmarkers, which formula is expressed by: 
\begin{equation}
    \label{eq:rep_corr}
    \frac{1}{S}\sum_{s=1}^{S}corr(
    {[d(\varphi(X^{(i, 1)}, y^{(i, 1)}), \varphi(X^{(i, 2)}, y^{(i, 2)}))\\]}_{i=1,\dots,N},
    {[d(l^{(i, 1)}, l^{(i, 2)})]}_{i=1,\dots,N}
    ).
\end{equation}
Additionally, for the approach producing landmarkers vector, we calculate the correlation of the distances between landmarkers and prediction with distances between landmarkers expressed by:
\begin{equation}
    \label{eq:land_corr}
    \frac{1}{S}\sum_{s=1}^{S}corr(
    {[d(\varphi(X^{(i, 1)}, y^{(i, 1)}), l^{(i, 2)})\\]}_{i=1,\dots,N},
    {[d(l^{(i, 1)}, l^{(i, 2)})]}_{i=1,\dots,N}
    ).
\end{equation}

\subsubsection{Warm-start}
\label{sec:warmstart}
The second evaluation approach is based on comparing the performance gain in the target meta-task -- warm-starting Bayesian HPO. For such validation, we perform Bayesian HPO on each dataset from the meta-validation sample. The optimization consists of $20$ iterations with $5$ warm-start configurations. We check the results after the initial phase and after the entire optimization. 

For all strategies based on encoders, we use the same hyperparameter sampling strategy.  Given representations of the datasets from the hyperparameter base and representation of the considered dataset, we search for the closest neighbors in terms of the Euclidean distance between representations. Having such datasets, we choose the best configuration from each of them and propose it as a warm-start point. For the gain evaluation, we use two metrics. The first is an ADTM plot with a scaled objective value, which shows the performance of HPO over time. The second is a CD plot, which is the result of the Friedman~\citep{friedman_test} and Wilcoxon-Holm~\citep{Holm1979ASS} tests, which show the statistical significance of the results.

\subsection{Baselines}
The process of evaluation consists of two types. The first approach is the correlation of the distances between representations produced by encoders and between landmarkers, and the second one is the evaluation of Bayesian HPO warm-start potential. In the case of the first method, as a baseline, we used basic Dataset2Vec from~\citep{dataset2vec}. For the second one, we include four additional:
\begin{enumerate}
    \item No warm-start - A random sampling of the configuration of hyperparameters \textbf{from entire hyperparameter space}, which can yield a reasonable lower bound of performance but does not result in a reliable comparison of selection from portfolio algorithms.
    \item Random from portfolio - A random selection of configurations \textbf{from portfolio} which can give a lower bound of performance of the selection algorithm.
    \item Rank - fixed selection of the best configurations on average, which shows how good the solution is based on a simple heuristic that is computationally cheap and easy to implement.
    \item Landmarkers - selection using nearest neighbors using true landmarkers. In a real-world setting, this approach is impractical since it requires a lot of models to be trained and evaluated. Still, this baseline can provide an upper bound on the performance of selection algorithms.
\end{enumerate}

\subsection{Results}

\subsubsection{Encoders training}
Each method is trained with respect to different objectives. In the case of Dataset2Vec, there is a cross-entropy loss in the classification of the origin of two subsamples of data. For metric learning and landmarker prediction, approaches are the loss functions described in Sections~\ref{sec:dataset2vec_metric} and~\ref{sec:dataset2vec_reconstruction}. As a result of the training of encoders, we obtain encoders with the following metrics:
\begin{itemize}
    \item Dataset2Vec encoder with validation accuracy of 0.85 in meta-task of classification whether two samples of data originate from the same dataset,
    \item landmarker reconstructor with reconstruction loss of 0.0307, which is lower than the initial value at the beginning of the training -- 0.0256,
    \item metric-based encoder with distance-based MSE of 0.0040, much lower than its initial value -- 0.0071.
\end{itemize}
These results show that all approaches can learn in terms of defined objectives.

\subsubsection{Distance correlation}
In Table~\ref{tab:correlations}, we show the correlation between the information produced by encoders (appropriate to the method) and landmarkers vectors described in Section~\ref{sec:distance_corr}. It can be seen that the best approach in terms of this metric is a metric-based approach, which significantly outperforms other encoders in terms of metric from Equation~\ref{eq:rep_corr}. On the other side, the highest correlation is achieved by the reconstruction approach using Equation~\ref{eq:land_corr}. In addition, all proposed approaches seem to learn some properties of landmarkers since the correlation coefficient, in most cases, is significantly higher than 0. Also, one can see that the basic version of Dataset2Vec, while performing satisfactorily in its original meta-task, yields no information contained within landmarkers.
\begin{table}[h]
    \centering
    \caption{Rank correlations between encoders' outputs and landmarkers}
    \label{tab:correlations}
    \begin{tabular}{lrr}
        \hline
Encoder                                                                                         & Correlation's average & Correlation's standard deviation \\ \hline
Dataset2Vec basic (Equation~\ref{eq:rep_corr})                                                                               & 0.037                 & 0.024                           \\ \hline
Dataset2Vec metric (Equation~\ref{eq:rep_corr})                                                                     & 0.332        & 0.030                 \\ \hline
Dataset2Vec reconstruction (Equation~\ref{eq:rep_corr}) & 0.177                & 0.027                            \\ \hline
Dataset2Vec reconstruction (Equation~\ref{eq:land_corr})                                                                     & 0.470                 & 0.028                            \\ \hline
\end{tabular}
\end{table}

\subsubsection{Warm-start}
\begin{figure}[h]
    \begin{subfigure}[t]{\textwidth}
        \centering
        \includegraphics[width=0.75\linewidth]{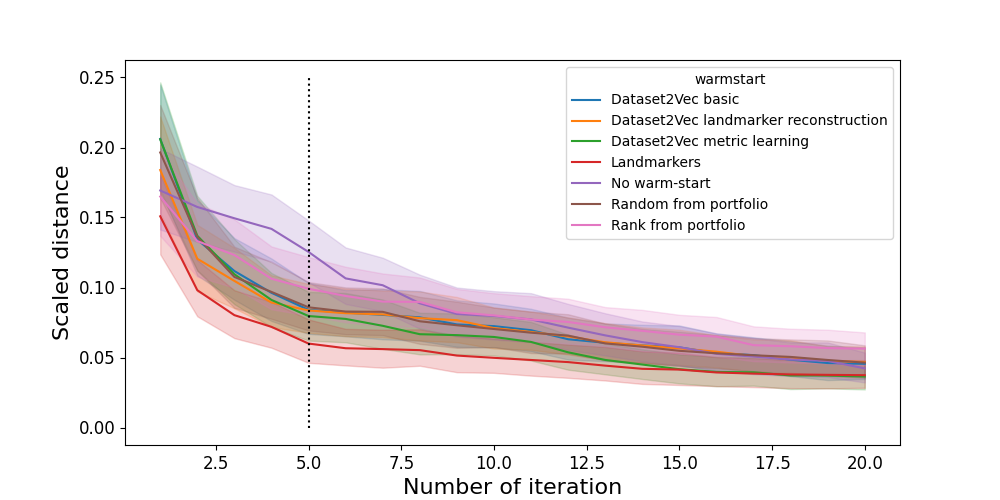}
        \caption{ADTM Plot}
        \label{fig:adtm}
    \end{subfigure}
    \hfill
    \centering
    \begin{subfigure}[b]{\textwidth}
        \centering
        \includegraphics[width=0.75\linewidth]{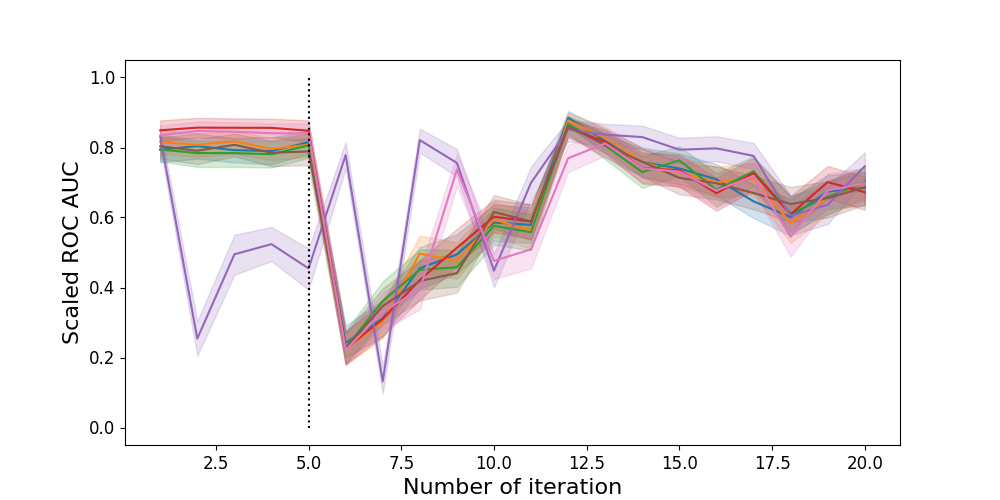}
        \caption{Scaled ROC AUC}
        \label{fig:raw_values}
    \end{subfigure}
    \caption{Comparison of performance between warm-start approaches. The dotted line indicates the moment when warm-start ends and optimization begins.}
    \label{fig:comparison}
\end{figure}

After training the encoders, we evaluate their usability in warm-starting Bayesian HPO. In Figure~\ref{fig:comparison}, we present different aspects of the performance of warm-start approaches. Figure~\ref{fig:adtm} shows an ADTM plot during optimizations comparing all warm-start methods. In Figure~\ref{fig:raw_values}, we extend that comparison showing scaled values of ROC AUC metric for candidate configuration in each iteration for all considered approaches. One can see that, in that case, the landmarker reconstruction is better than metric learning. Additionally, it can be seen that optimization, regardless of the approach to the warm-start, has a significant decrease in the objective function after the initial phase, which can indicate that the optimizer has too excessive exploration factor. 

\begin{figure}[h]
     \centering
     \begin{subfigure}[b]{0.9\textwidth}
         \centering
         \includegraphics[width=\textwidth]{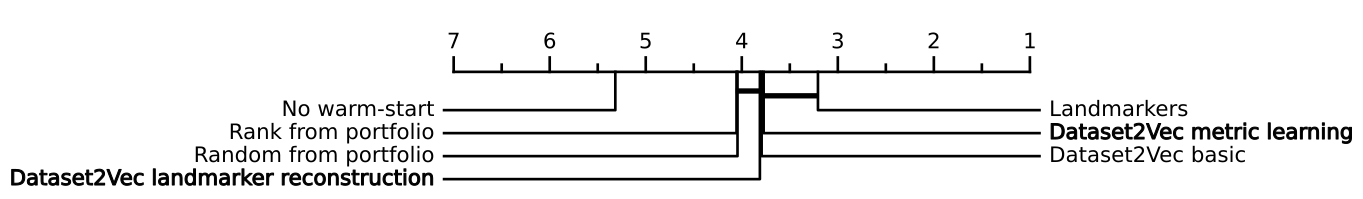}
         \caption{Comparison of objective values at the end of the warm-start phase.}
         \label{img:cd_5}
     \end{subfigure}
     \vfill
     \begin{subfigure}[b]{0.9\textwidth}
         \centering
         \includegraphics[width=\textwidth]{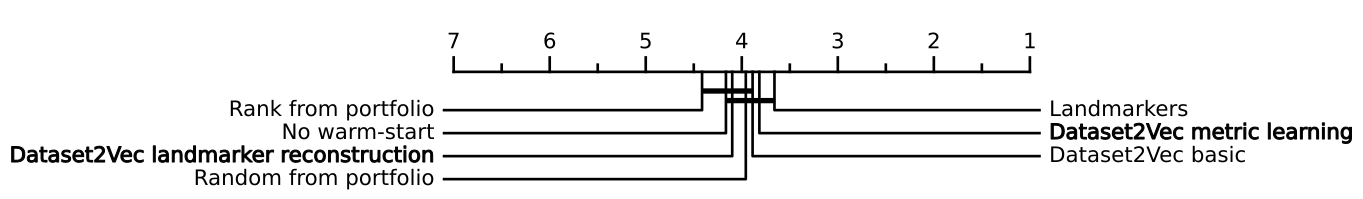}
         \caption{Comparison of objective values at the end of optimization.}
         \label{img:cd_20}
     \end{subfigure}
     \caption{Comparison of warm-start approaches on CD plots. The more the vertical bar is to the right, the better. Approaches connected with the horizontal bar are statistically indistinguishable regarding the Wilcoxon-Holm test.}
     \label{img:cd_plots}
\end{figure}

Figure~\ref{img:cd_plots} shows CD plots summarizing the significance of differences in the effectiveness of all approaches at two points: just after warm-start (Panel a) and after all iterations of Bayesian HPO (Panel b). 
What is important, the results after the initial phase confirm that landmarkers method is the best and statistically superior to baselines using random warm-start points. So, we can conclude that \hyperref[ass:land_align]{Requirement} is justified.  Encoder-based approaches outperform baseline selection algorithms but are statistically indistinguishable from other portfolio-based methods. At the same time, the proposed methods are not statistically worse than the upper bound represented here by landmarkers approach. \textbf{This means that representations produced by encoders can partially substitute landmarkers in the warm-start meta-task.}
Looking only at the encoders, the best one is based on metric learning. However, after all iterations of optimization, the differences between solutions become even smaller, and even the best landmarker-based method is not statistically different from the two baseline approaches: no warm-start and random sampling from the portfolio. We can conclude that Bayesian HPO benefits from warm-start points in a limited way, and their selection is crucial, especially to maximize anytime performance.

%% file: sections/discussion.tex
\section{Discussion}
The results show that the encoders manage to capture landmarkers' dependencies of dataset structure. This can be seen in Table~\ref{tab:correlations}, which shows how much distances between representations correlate with distances between landmarkers. On the other hand, we show that such representations yield insignificant gains in the target meta-task. It can have two causes. The first one is too weak information about the landmarkers that are yielded by encoders. The second cause is the fact that the Bayesian HPO does not benefit from the specific warm-start algorithm but the set of configurations from the specific hyperparameter subspace. On the other hand, we show that after the initial phase, the performance of the configurations significantly decreases. Additionally, after the entire optimization, it can be noticed that there is no significantly the best approach to the optimization warm-start. Therefore, the Bayesian HPO may be, to some extent, robust to the selection of the warm-start configurations at the beginning of the procedure. Hence, there is a need to refine the definition of meta-task since vanilla Bayesian Optimization methods are limitedly affected by warm-start points.

Our results show that the datasets' encoders are able to produce representations that align with landmarkers, but they are insufficient to be used in real-world applications. Further research in this field may consist of novel approaches to the architecture of the encoders, which may yield stronger information about landmarkers. The other branch of research that needs to be explored is novel metrics of the evaluation of the encoders. In this work, we proposed one. However, in our opinion, it is not sufficient. Additionally, a more detailed analysis of the impact of the warm-start itself on the entire optimization procedure should be performed.
\newpage

\noindent
\textbf{Social impact statement}

\noindent
After careful reflection, we have determined that this work presents no notable negative impacts on society or the environment.

%% file: sections/checklist.tex
\newpage
\section*{Submission Checklist}


\begin{enumerate}
\item For all authors\dots
  \begin{enumerate}
  \item Do the main claims made in the abstract and introduction accurately
    reflect the paper's contributions and scope?
    \answerYes{}
  \item Did you describe the limitations of your work?
    \answerYes{}
  \item Did you discuss any potential negative societal impacts of your work?
    \answerYes{}
  \item Did you read the ethics review guidelines and ensure that your paper
    conforms to them? (see \url{https://2022.automl.cc/ethics-accessibility/})
    \answerYes{}
  \end{enumerate}
\item If you ran experiments\dots
  \begin{enumerate}
  \item Did you use the same evaluation protocol for all methods being compared (e.g.,
    same benchmarks, data (sub)sets, available resources, etc.)?
    \answerYes{}
  \item Did you specify all the necessary details of your evaluation (e.g., data splits,
    pre-processing, search spaces, hyperparameter tuning details and results, etc.)?
    \answerYes{}
  \item Did you repeat your experiments (e.g., across multiple random seeds or
    splits) to account for the impact of randomness in your methods or data?
    \answerYes{}
  \item Did you report the uncertainty of your results (e.g., the standard error
    across random seeds or splits)?
    \answerYes{}
  \item Did you report the statistical significance of your results?
    \answerYes{}
  \item Did you use enough repetitions, datasets, and/or benchmarks to support
    your claims?
    \answerYes{}
  \item Did you compare performance over time and describe how you selected the
    maximum runtime?
    \answerYes{}
  \item Did you include the total amount of compute and the type of resources
    used (e.g., type of \textsc{gpu}s, internal cluster, or cloud provider)?
    \answerYes{}
  \item Did you run ablation studies to assess the impact of different
    components of your approach?
    \answerNA{}
  \end{enumerate}
\item With respect to the code used to obtain your results\dots
  \begin{enumerate}
\item Did you include the code, data, and instructions needed to reproduce the
    main experimental results, including all dependencies (e.g.,
    \texttt{requirements.txt} with explicit versions), random seeds, an instructive
    \texttt{README} with installation instructions, and execution commands
    (either in the supplemental material or as a \textsc{url})?
    \answerYes{}
  \item Did you include a minimal example to replicate results on a small subset
    of the experiments or on toy data?
    \answerYes{}
  \item Did you ensure sufficient code quality and documentation so that someone else
    can execute and understand your code?
    \answerYes{}
  \item Did you include the raw results of running your experiments with the given
    code, data, and instructions?
    \answerYes{}
  \item Did you include the code, additional data, and instructions needed to generate
    the figures and tables in your paper based on the raw results?
    \answerYes{}
  \end{enumerate}
\item If you used existing assets (e.g., code, data, models)\dots
  \begin{enumerate}
  \item Did you cite the creators of used assets?
    \answerYes{}
  \item Did you discuss whether and how consent was obtained from people whose
    data you're using/curating if the license requires it?
    \answerNA{}
  \item Did you discuss whether the data you are using/curating contains
    personally identifiable information or offensive content?
    \answerNA{}
  \end{enumerate}
\item If you created/released new assets (e.g., code, data, models)\dots
  \begin{enumerate}
    \item Did you mention the license of the new assets (e.g., as part of your
    code submission)?
    \answerYes{}
    \item Did you include the new assets either in the supplemental material or as
    a \textsc{url} (to, e.g., GitHub or Hugging Face)?
    \answerYes{}
  \end{enumerate}
\item If you used crowdsourcing or conducted research with human subjects\dots
  \begin{enumerate}
  \item Did you include the full text of instructions given to participants and
    screenshots, if applicable?
    \answerNA{}
  \item Did you describe any potential participant risks, with links to
    institutional review board (\textsc{irb}) approvals, if applicable?
    \answerNA{}
  \item Did you include the estimated hourly wage paid to participants and the
    total amount spent on participant compensation?
    \answerNA{}
  \end{enumerate}
\item If you included theoretical results\dots
  \begin{enumerate}
  \item Did you state the full set of assumptions of all theoretical results?
    \answerNA{}
  \item Did you include complete proofs of all theoretical results?
    \answerNA{}
  \end{enumerate}
\end{enumerate}

%% file: sections/appendix.tex
\section{Hyperparameters grid}
\label{app:hp_grid}

\begin{table}[h]
\centering
\caption{Hyperparameters grid of the considered xgboost algorithm.}
\begin{tabular}{lrr}
\hline
Hyperparameter     & Range               & Distribution         \\ \hline
n\_estimators      & {[}10, 1000{]}      & Uniform (int)        \\ \hline
eta                & {[}0.00001, 1{]}    & Uniform (float, log) \\ \hline
gamma              & {[}0.00001, 1{]}    & Uniform (float, log) \\ \hline
max\_depth         & {[}3, 8{]}          & Uniform (int)        \\ \hline
min\_child\_weight & {[}0.00001, 100{]}  & Uniform (float, log) \\ \hline
reg\_lambda        & {[}0.00001, 1000{]} & Uniform (float, log) \\ \hline
reg\_alpha         & {[}0.00001, 1000{]} & Uniform (float, log) \\ \hline
\end{tabular}
\end{table}